\documentclass{article}

\PassOptionsToPackage{numbers, compress}{natbib}
     \usepackage[preprint]{neurips_2020}

\usepackage[utf8]{inputenc} 
\usepackage[T1]{fontenc}    
\usepackage{hyperref}       
\usepackage{url}            
\usepackage{booktabs}       
\usepackage{amsfonts}       
\usepackage{nicefrac}       
\usepackage{microtype}      

\usepackage{graphicx}
\usepackage{subfigure}
\usepackage{amsthm}
\usepackage{amsmath}
\usepackage{algorithm}
\usepackage{algorithmic}

\theoremstyle{definition}
\newtheorem{definition}{Definition}[section]
\usepackage{caption}
\title{Neural Networks and Denotation}

\author{%
  Eric E. Allen \\ 
  Acrisure Technology Group\\
  Austin TX \\
  \texttt{eric@acrisuretechnology.com} \\
}


\begin{document}

\maketitle

\begin{abstract}
We introduce a framework for reasoning about what meaning is captured by the neurons in a trained neural network.  We provide a strategy for discovering meaning by training a second model (referred to as an \emph{observer model}) to classify the state of the model it observes (an \emph{object model}) in relation to attributes of the underlying dataset. We implement and evaluate observer models in the context of a specific set of classification problems, employ heat maps for visualizing the relevance of components of an object model in the context of linear observer models, and use these visualizations to extract insights about the manner in which neural networks identify salient characteristics of their inputs. We identify important properties captured
decisively in trained neural networks; some of these properties are denoted by individual neurons. 
Finally, we observe
that the label proportion of a property denoted by a neuron is dependent on the depth of a neuron within 
a network; we analyze these dependencies, and provide an interpretation of them. 
\end{abstract}

\section{Introduction}
\label{introduction}

Artificial neural networks have come to dominate the field of machine learning in a variety of contexts, including supervised learning and reinforcement learning. Despite their explanatory power, a key drawback of neural networks lies in the difficulty of understanding what they have learned; this knowledge is locked in the weights of a network's layers. There are many ways to approach the problem of discovering what a neural network learns. One thread of research has involved explaining the predictions of a network in relation to features of the data. We can attempt to capture such explanations by weighing the importance of specific features \citep{lundberg2017unified} or by sending combinations of features through linear and non-linear transformations \citep{vaughan2018explainable}. Alternatively, we can perturb features and measure the impact \citep{ribeiro2018anchors}. But a common property of these approaches is that they focus directly on the data provided to the network. In contrast, we propose to view a network itself as a representation of learned knowledge and ask: How does a neural network \emph{denote} concepts that are relevant to, but distinct from, the property it is has been trained to identify?

Traditionally, when neural networks are manually inspected, an emphasis is placed on deciding what specific neurons (or collections of neurons) denote \citep{ActivationAtlases}. Some of the insights gleaned by such an approach can be quite powerful. However, this technique relies on time consuming human inspection of a learned network, and the patterns found are inherently subjective. Moreover, what if some concepts do not correspond directly to the activation of specific neurons? It is conceptually possible that properties of the data are denoted by context-dependent patterns in the firing of neurons throughout a network. For example, what if a group of neurons in a convolutional network for video acts as a detector of some property of a physical object's shape (for example, roundedness) when an input image is well-lit, but reverts to an entirely different function in low light environments (such as simple movement)? Could multiple groups of neurons perform the same function with varying competencies in distinct contexts? What if the very identities of the groups of neurons in a network that denote salient properties are dependent on the input? Consider, for example a reinforcement learning agent trained to play Pac-Man \citep{Atari57}. We might find that a collection of neurons maps the distance to each ghost in a game of Pac-Man so long as Pac-Man himself is being chased, but the constituent neurons are drafted into other groups when Pac-Man is doing the chasing. And we can expect that some concepts are not explicitly denoted by collections of neurons at all, but are nevertheless logically entailed by the activations of neurons. For example, if a neural network encodes a full representation of an agent's environment, that representation entails many concepts (such as the paths available to the agent) that might not be represented explicitly. 

The space of possibilities concerning how the components of a neural network denote concepts is enormous. Indeed, the relationships between denotations of components could be so complex as to defy simple articulation. We might even question whether it is sensible to view sub-components of a network (in any context) as denoting anything at all. In order to explore questions concerning neural networks and denotation, we need a way to ascertain meaning from neural networks in a manner that makes as few assumptions as possible about how that meaning is stored. We start with the question as to whether meaning can be found at all. By making this question precise, and introducing a  framework to study the question, we are able to put important constraints on the space of possible answers. 

Even with a precise understanding of how neurons might denote specific concepts, we are still faced with the problem of identifying such denotations in a trained network. As it turns out, we have a powerful tool available to detect meaning in structure: Neural networks themselves. Therefore we devise the following approach: We train a network as we normally would in a supervised setting; we call this trained network an \emph{object model}. We then use the neural activations of the object model, on an appropriate dataset, as a new feature set. We choose a set of salient properties of our dataset that are distinct from the labels used for object model training, and use each of these salient properties as \emph{labels} on a new feature set. For each of these properties, we then train a separate neural network, called an \emph{observer model}, to predict the label given only snapshots of the neural activations of the object model on the corresponding input.
\footnote{We like to think of these snapshots of the object network as analogous to fMRIs of biological neural networks.}

Observer models are a powerful framework in which to explore many questions about the nature of denotation in a neural network. Most important, they help us to determine whether certain concepts are captured by a trained network at all. But they also provide us with a tool for exploring more nuanced questions we have concerning denotation in a neural network. For example, by exploring the required complexity of an observer model, we form insights into the way in which neurons must be capturing a concept in our trained object model. 
Moreover, by masking and transforming aspects of the activations provided to an observer model and measuring how such alterations affect performance, we can extract additional insights regarding the manner in which meaning is stored. In this paper, we demonstrate that important properties are denoted even by individual 
neurons in a network. We also identify quantifiable aspects of the types of properties denoted by 
neurons that is dependent on their location within a deep network structure, and we provide an interpretation
of this dependency. 

\section{Preliminaries}
\label{theory}
Although the central ideas of our approach are intuitively clear, we have found it helpful to formalize
the notion of neural denotation, so as to clarify our thinking, ensure precise meaning of terms,
and avoid numerous missteps in the development of this work. 

\subsection{Tensors as data, models, and activations}
Conveniently, we use tensors to capture the structure of all three central subjects of our study: Data, model parameters, and model activations.
We view an $n$-tensor as a mapping from all $n$-tuples in a range of natural numbers 
(with an upper bound along each axis)
to real scalar values. We call each $n$-tuple in the domain of a tensor a \emph{position}. 
We call a subset of the positions in the domain of a tensor a \emph{silhouette}. 
The \emph{shape} of a tensor is the sequence of upper bounds of its axes.
We use the term \emph{vector} synonymously with \emph{1-tensor} and \emph{matrix} synonymously with \emph{2-tensor}.

Let $D$ be a \emph{dataset} $D = d_1, ..., d_m$ consisting of $m$ \emph{data points}, where each data point $d_k$ is a pair $X_k, y_k$ consisting of an $n$-tensor $X_k$ of fixed shape (called a \emph{feature tensor}), and a label vector $y_k$ of fixed length. A dataset is \emph{fit} by passing it to a 
non-deterministic function 
$F$, which maps $D$ to a \emph{model tensor} $M$ denoting the parameters of a trained model. For example, the parameters of a fully connected neural network denote the weights of a trained network, arranged in a 3-tensor, where each slice along the primary axis denotes the matrix of weights for a single layer; multilayer perceptrons with layers of
varying width are modeled by padding narrow layers with zeros.  Given a model tensor $M$ and a feature tensor $X_k$, we \emph{predict} a label for $X_k$ via a pair of functions: An \emph{activation} operator $G$, mapping the pair $M, X_k$ to an \emph{activation tensor} $A_k$, and a \emph{predictor} function $P$, mapping ${A_k}$ to a label vector ${{y}_{A_k}}$. Conceptually, an activation tensor denotes a model's response to a given stimulus; in the case of a fully connected network, it denotes a matrix of neural activations such that each row denotes a layer of activations. The \emph{performance} of $M$ is a scalar measure of the correspondence of predictions ${{y}_{A_k}}$ with labels $y_k$. The best choice of performance measure depends on context; in our 
formalism,  
we leave it abstract.

We extend the operators $G$ and $P$ to datasets in a straightforward manner: Given a dataset $D$, we write  $G(M, D)$  to refer to the image of $D$ under $G$ with parameters $M$. Given a sequence of activations $A$, we write $P(A)$ to refer to the image of $A$ under $P$. 

\subsection{Neural activations as sensory data}
The notion of  all ``properties'' we might consider to be inherent in a feature tensor is ambiguous. We make this notion precise by following the approach taken in denotational semantics and identify properties with their extension \citep{ebbinghaus2013mathematical, scott1982domains}.
Given the population $\Phi$ of all feature tensors expressible to a given model, a \emph{property} 
is a subset $\phi$ of feature tensors in $\Phi$. The elements $X$ in $\phi$ are said to \emph{possess} the property $\phi$. 

We can map a property $\phi$ to a collection $G(M, {\phi})$ of activation tensors. We are interested in exploring whether there is a common pattern providing evidence for $\phi$
in these activation tensors; if so, we would like to view that pattern as \emph{denoting} the property $\phi$. 
How can we tell if such a pattern exists? Consider that the collection of all activation tensors $G(M, \Phi)$ itself can be thought of as a collection of feature tensors. We can label the elements in this collection according to whether they are in $G(M, \phi)$. We propose to identify the existence of a pattern
in the activation tensors $A$ by training a model on our new collection and determining whether the performance of the model exceeds a performance threshold.  

\theoremstyle{definition}
\begin{definition}[Denotation]
Let $D = d_0, ..., d_m$ be a dataset, $\phi$ be an auxilliary property of the feature tensors in $D$, 
$M$ a model in $F(D)$, and $S$ a silhouette 
of positions in $G(M, D)$. Construct a set of feature vectors consisting of $G(M, D)$ restricted to the positions in $S$. For each $d_k = (X_k, y_k)$ in $D$, label $G(M, d_k)$ according to whether $X_k \in \phi$. 
Call the resulting dataset $E$. If there exists a model ${O}$ in ${F(E)}$ and a threshold $t$ such that the performance of ${O}$ exceeds $t$, then we say that the silhouette $S$ \emph{denotes} $\phi$ with threshold $t$.
We call $M$ an \emph{object model} and $O$ an \emph{observer model}.
\end{definition}

Note that a single silhouette might denote multiple properties, each with its own threshold performance. Also, there can be multiple denotations of $A$ in a single model; some of these collections might be contained in others; some pairs of denotations might overlap; some might be disjoint. 

\subsection{Types of observer models}
A particularly straightforward observer model is an \texttt{and}-gate that identifies whether 
every neuron in the positions in its input 
silhouette have activated (i.e., have an activation value greater than a fixed threshold value, such as
zero in the case of ReLU activation). We refer to this case as
\emph{simple denotation}. This form of denotation corresponds to what we might intuitively think of
as denotation in a neural network when manually inspecting the activations.  

To take one step beyond simple denotation, consider the case where our observer model is a linear classifier, 
for example a logistic regression classifier, support-vector machine, or a single-layer perceptron as described by Rosenblatt \citep{theil1969multinomial, shawe2000support, rosenblatt1957perceptron, minsky2017perceptrons}. 
In this case, the contribution of a silhouette $S$ of neural activations is determined solely by the weights assigned to each component of $S$ (some of which might be negative) together with a bias vector. We call this case
\emph{linear denotation}. 

As our discussion of conditional denotation indicates, it is entirely possible that threshold performance for a label class $z$ cannot be achieved with a linear model. Indeed, one of the issues we wish to explore is how often a linear model \emph{is} sufficient for denotation. When it is not sufficient, we might find that rich network architectures, such as multilayer perceptrons and convolutional networks, are able to achieve threshold performance. In such cases, we can infer that ({\it{i}}) the property we wish to identify is captured in the activations of the observer model but ({\it{ii}}) there is no silhouette of neural positions that can be said to denote the property linearly. In such cases, we might be able to uncover linear denotation by building a chain of linear observer models, starting by discovering linear denotations of simpler properties in the object model, and 
extending the chain by training each subsequent model to observe increasingly complex properties in the activations of the observer model before it. 

\begin{figure}[t!]
\begin{center}
\begin{tabular}{@{}lr@{}} \toprule
Model & Label proportion\\ \midrule
Material Advantage & 0.76 \\
White in Check & 0.047 \\
White has insufficient material to win & 0.0006 \\ \bottomrule
\end{tabular}
\caption{Label proportions for three observer datasets: Checking whether white has material advantage, whether white is in check, and whether white has sufficient material to win. The labels of each dataset are biased, as are these three properties in a distribution of human chess games.}
\vspace{-6mm}
\label{biases}
\end{center}
\end{figure}

\section{Experimental Design}
\label{approach}
In order to explore observer models experimentally, we focus our attention on object models trained on collections of chess boards. We choose to explore this question in the context of chess boards because ({\it{i}}) there is a wealth of example boards in the chess literature, labeled with established moves and concepts, ({\it{ii}}) it is straightforward to programmatically annotate chess boards with additional labels, and ({\it{iii}}) there are well-known approaches and architectures for applying machine learning to chess \citep{AlphaZero, campbell2002deep}. 

To construct an object model, we train a network to look at a chess board and choose 
which piece to move, using a labeled dataset from human games. We make use of the DeepIntuit open source chess engine and datasets \citep{DeepIntuit}.
Our approach for representing boards and predicting best moves builds on the approach taken by DeepIntuit and described in \citep{oshri2016predicting}.
Boards are denoted by $8\times8\times6$ tensors. Each of the six layers denotes the positions of a distinct type of piece: pawns, knights, bishops, rooks, queens, kings. The positions of white pieces are denoted with $1$, black with $-1$, and empty squares with 0. A large corpus of boards taken from expert human games, labeled with the position of the next piece moved, are used as a training set. All boards are normalized so that the next player to move is white; this is done by taking boards where black is to move, reflecting positions vertically along the center of the board, and inverting the colors. While this might have undesirable consequences for agents that plan long sequences of moves, the negative impact should be minimal when simply predicting the next piece to move. In total, our training set consists of $633,586$ labeled boards. We keep a test set of 211,196 boards. 

The object model we consider is a multilayer perceptron, consisting of three layers, with $128$ neurons per layer, and ReLU activation. The output layer is a softmax layer with $64$ neurons, indicating the position of the piece predicted to move on the next turn. After each layer, a \emph{recording} layer is inserted, enabling us to take a snapshot of the activations of the layer before it. All models are built using the Tensorflow Keras library \citep{tensorflow2015-whitepaper}. 

Predicting which piece to move with a static model is far from state of the art, and is of course  na\"ive when compared to modern systems such as AlphaZero \citep{AlphaZero}. Nevertheless, such an object model serves our current purpose well; all properties of the board learned by our model must be directly captured in the parameters of the network itself. Contrast with approaches such as DQN or MCTS where important experience is captured in auxilliary data structures \citep{sorokin2015deep, browne2012survey}. 
We employ early stopping and use 20\% of training data for validation. The validation set is randomly selected at the start of each training run. Training proceeds using Adam optimization, 
categorical cross-entropy as loss, and a batch size of 128. Final accuracy was 40\% on training data and 37\% on test data, matching reported results on this dataset \citep{DeepIntuit}.

\section{Implementing Observer Models}
\label{explanatory}

After training our object model, we then build a new set of feature tensors by taking snapshots of the 
hidden layers of our network when looking at single boards in our test set. 
We generate three distinct sets of labels for those same boards by computing ({\it{i}}) which player has a material advantage (black or white), according to the standard weightings of chess pieces values, ({\it{ii}}) whether the white player is in check and ({\it{iii}}) whether the white player still has sufficient material to win. For example, if white has no pieces other than a king and bishop, a win is impossible. We label the corresponding activations with each of these label sets, forming three observer datasets. Note that these labels have no direct relation to the labels that the object model was trained on; instead, they denote interesting attributes of a board that (we hope) can be gleaned from our activations of the object models.

Finally, we train the following observer models on each of these new datasets:
\begin{itemize}
\item Logistic regression. We use this model to detect linear denotation. 
\item A multilayer perceptron consisting of three dense hidden layers with $256$ neurons each, employing ReLU activation, and a single output neuron with sigmoid activation. 
\item A convolutional network with three convolutional layers employing $3\times3$ kernel size, same padding, and $32$ channels, followed by two dense layers consisting of $256$ neurons each, and finally a single output neuron with sigmoid activation. The convolutional observer model views the object model as possessing a two-dimensional geometry that respects its matrix of activations. 
\end{itemize}
As with our object model, we employ Adam optimization, 
categorical cross-entropy as loss, and a batch size of 128. We reserve 20\% of the dataset for validation, 
again randomly selected during training. Finally, we construct an observer test set by running our original test set through the object model, recording the activations, and labeling appropriately. 

The properties of being in check, and of possessing sufficient material to win, are highly biased in the dataset (as they are in most chess games); we report label proportions in Figure \ref{biases}. To control for biased labels, we report both accuracy and F1 scores for our observer models. 

\begin{figure}[t!]
\begin{center}
\begin{tabular}{@{}llllr@{}} \toprule
Model & Train Accuracy & Test Accuracy & Train F1 Score & Test F1 Score\\ \midrule
Material Advantage (Linear) & 0.77 & 0.77 & 0.86 & 0.86 \\
Material Advantage (MLP) & 0.86 & 0.81 & 0.86 & 0.86 \\
Material Advantage (Conv) & 0.99 & 0.81 & 0.87 & 0.87 \\
White in Check (Linear) & 0.95 & 0.95 & 0.09 & 0.09 \\
White in Check (MLP) & 0.96 & 0.95 & 0.10 & 0.10 \\
White in Check (Conv) & 0.99 & 0.95 & 0.16 & 0.15 \\
Sufficient to Win (Linear) & 0.99 & 0.99 & 0.0014 & 0.0014 \\
Sufficient to Win (MLP) & 0.99 & 0.99 & 0.0013 & 0.0012 \\
Sufficient to Win (Conv) & 1.00 & 0.99 & 0.0584 & 0.1310 \\ \bottomrule
\end{tabular}
\caption{Accuracies and F1 Scores on training and test data for a variety of observer models on an object model that chooses positions of chess pieces to move. We use three observer labels: Checking whether white has material advantage, whether white is in check, and whether white has sufficient material to win. Each label type is tested on three distinct neural architectures: A linear model (logistic regression), a multilayer perceptron, and a convolutional network.}
\vspace{-5mm}
\label{performance}
\end{center}
\end{figure}

\section{Results}
\label{results}

The results of our observer models on each dataset are shown in Figure \ref{performance}. Material advantage is predictable with high accuracy, even by a linear model. The other two properties are not well predicted, as their F1 scores indicate. However, all three properties are best predicted by a convolutional observer model, 
by as much as two orders of magnitude when predicting sufficient material to win.  

Our experiments demonstrate that it is possible to train a model to infer salient properties of a dataset by viewing the activations of a neural network examining it. Moreover, by measuring the complexity of the explanatory model needed to elicit these attributes, we shed light on the manner in which the attributes must be stored in the underlying network. A linear model can achieve high accuracy in denoting attributes only if those attributes are captured directly in the activations of the object model. 
Our results suggest that many aspects of salient properties are present in the object model but require more complex observer models to discover. 

\begin{figure}[t!]
\vskip 0.2in
\begin{center}
\scalebox{1}[1]{\centerline{\includegraphics{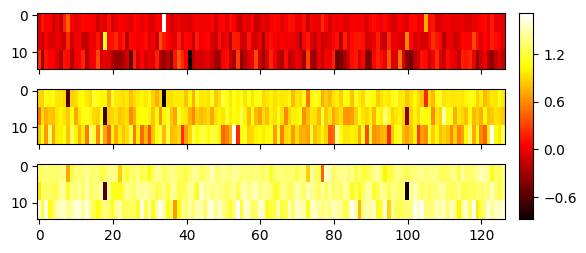}}}
\vspace{-5mm}
\caption{Heat maps of linear observer models for three properties. Top: for determining material advantage. Middle: For determining whether the white king is in check. Bottom: for determining whether white has sufficient material to win. The weights for each heat map are arranged to mirror the neural architecture of the object model: A fully connected network of three layers with 128 neurons each. Layers closer to the input are lower.}
\vspace{-2mm}
\label{many_linear_heat_maps}
\end{center}
\vskip -0.2in
\end{figure}

\section{Neural heat maps}
We would like to gain intuition concerning what a linear observer model learns about the parameters of an object model. One way to do this is to construct a heat map
of the weights that the observer model assigns to the parameters of the object model. Because the contributions of the object model's parameters are determined linearly in this case, a heat map shows us directly how much each object model parameter is contributing to an observer model's inference.\footnote{Both strongly positive and strongly negative values in a heat map indicate neurons that contribute to the observer model significantly. We leave weights as signed values so as to distinguish these two cases in the visualization.}  Heat
maps for the linear observer models concerning material advantage, check, and possessing sufficient material to win are shown in Figure \ref{many_linear_heat_maps}. The weights from each object model neuron are arranged graphically so as to mirror the architecture of the object model itself: A three layer multilayer perceptron with 128 neurons at each layer. Because we are interested only in the relative contributions of each neural activation, we do not display bias vectors. As our heat maps indicate, the weights assigned to the object model parameters span a broad range. More heavily weighted parameters do not cluster, and contributions to the observer model output occur throughout the object model. 

Intriguingly, in the case of the material advantage observer model, significant weight is placed on two neural activations in the second and third layers, with relatively small weights on all other neurons. These weights suggest that the silhouette of these two neurons alone might denote the property of material advantage with high threshold.  
To explore this possibility, we construct an alternative observer dataset consisting only of these two neural activations, annotated with our material advantage labels, and feed the resulting dataset to a logistic regression model. If logistic regression can classify material advantage based solely on these two neurons, 
then there must be a threshold value associated with at least one of these neurons that denotes the property that white has material advantage. We also construct two more datasets, each using the activations of just one of the two neurons in the silhouette. 

Running logistic regression on this silhouette results in an F1 score matching that of our linear observer model trained on the full set of activations. {\emph{Moreover, the F1 score of an observer trained on either of the two neurons individually matches that of the observer trained on all activations}}. 
Given this result, we conclude there are at least two examples of a single neuron in our object network denoting material advantage with threshold $0.86$. This satisfying result demonstrates the value of using heat maps in concert with the framework we have devised to spot denotation in neural networks. 

\section{Label proportions of neuron-denoted properties}
\label{neuron-denoted}

\begin{figure}[t!]
\vskip 0.2in
\begin{center}
\scalebox{1}[1]{\centerline{\includegraphics{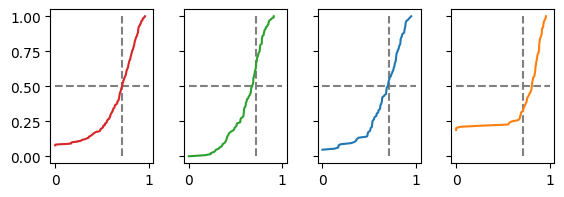}}}
\vspace{-5mm}
\caption{Cumulative distribution functions for label proportions of properties on the object model test set, 
denoted by neurons at each layer. 
First shown is all layers (red), followed by the first (green), second (blue), and third (orange) layers. 
A dashed vertical line is shown at the median label proportion for all neurons: 0.716.
Deeper layers include a greater proportion of properties with label proportions of zero. Label proportions for
each neuron on the training set match that of the test set to within 0.5\%.}
\label{neural_bias_cdfs}
\vspace{-2mm}
\end{center}
\vskip -0.2in
\end{figure}

We make two observations concerning our observer model results:
\begin{itemize}
\item Silhouettes consisting of individual neurons can denote important properties.
\item Of the three properties used for observer models, the property we successfully mapped to a silhouette had the least extreme label bias of the three.
\end{itemize}
These observations suggest another experiment. If we limit ourselves to silhouettes consisting of single neurons, and consider only simple denotation, it is easy to find the extension of the property denoted by an individual neuron: we simply feed our dataset to the object model and observe which data points result in activation of the neuron. For each neuron, we then ask: On what proportion of the dataset does the neuron activate? 
Because our object model uses ReLU activation, we consider a neuron to have activated if its activation value is greater
than zero. The results of this experiment are shown in Figure \ref{neural_bias_cdfs}. We graph cumulative distribution functions (cdfs) for the label 
proportions of neurons in our object model, for the entire network and for each layer. Label proportions for 
each neuron on the training and test sets match to within 0.005; thus the training and test cdf graphs are virtually indistinguishable. We show results generated from the test set. 

Our restriction to simple denotation is what enables us to ask what the label proportions are of the properties denoted
by each neuron. If we were to instead use linear denotation, we would be forced to choose a bias for each neuron, 
which would impact the very label proportions we are trying to measure. 

Three aspects of these cdfs stand out: ({\it{i}}) Most neurons in the object model denote properties with label proportions greater than 0.7. ({\it{ii}}) The median label proportion increases with deeper layers. ({\it{iii}}) The proportion of neurons denoting properties with no positive examples in the dataset (which we call \emph{annihilated neurons}) increases in deeper layers. 
All three aspects invite deeper investigation. We provide some  
suggestions for interpreting these results. 

The gradual increase in annihilated neurons at each subsequent layer effectively serves as a series of 
projections of the input into progressively lower dimensional spaces, i.e., the dimensions corresponding to the 
activations of the surviving neurons. Given that the dimensionality of the input layer of our object model
is much greater than that of the output layer (as it is in most deep learning models), 
it is reasonable to expect the network to learn to progressively project the input
into lower dimensional spaces as it processes the input toward a final output signal, acting as a sort of 
funnel. 
We might also wonder whether the 
progressive increase in annihilated neurons explains the gradual increase in label proportions with deeper 
layers. We test this by randomly annihilating a selection of neurons from the first layer so 
that the proportion of annihilated neurons matches that of the second layer, and then the third layer, and
checking whether the cdfs match what is observed. They do not; the median label proportions we observe in 
our experiment are $0.67$ and $0.63$ for the second and third layers respectively. 

If we view the vector of neural activations at a layer as a code compressing the input, the gradual increase 
in annihilated neurons implies that the degree of compression increases with deeper layers. For a given number
of segments we might wish to distinguish in our dataset, the Kraft inequality provides
an information theoretic lower bound on the number of active neurons needed to distinguish these segments, even if the denoted 
properties provide high information gain
\citep{kraft1949device}. But the observed high median label proportion, as well as the increase in label proportions with deeper layers, suggest that each neuron is providing low information gain to begin with, and
even less in deeper layers.
We might have expected the label proportions for properties denoted by neurons to cluster at 0.5, so that neural signals tend to maximize entropy reduction, a typical splitting strategy in decision tree algorithms such as ID3 \citep{quinlan1986induction}. But as the data indicate, such splits are relatively rare. In fact, only three neurons 
denote properties with label proportions between 0.49 and 0.51.  
Both the information gain per neuron, and the total number of neurons available for encoding a segment, is decreasing with deeper layers. Effectively, the network is losing a portion of its theoretical 
capacity to 
encode information with each new layer. How can we explain this phenomenon?

One advantage of a property with high label proportion is that the complement of the denoted property more precisely matches a smaller segment
of the data. Effectively, it seems as if 
the network is becoming increasingly adept at identifying niche segments in deeper layers, by
sacrificing the information capacity needed to distinguish a large swath of the population of expressible inputs. 
For real-world datasets, it is plausible that this tradeoff is worthwhile. For example, there are a large
number of states of chess boards that simply do not arise, and in fact it is impossible for there to be more than 32
pieces on a legal board. It would be reasonable for the network
to avoid using information capacity to encode unlikely or impossible inputs. 
Lower label proportions also match smaller data segments, but driving neurons to low (but non-zero) label proportions likely conflicts with the pressure to increase the set of annihilated neurons in deeper layers. 
It is straightforward for subsequent layers to invert label proportions of their inputs when needed, using negative weights.

Neural networks are known to have the capacity to 
memorize random data, and they behave differently on random data \citep{arpit2017closer}. One way to explore our interpretation of label proportions would be to train a network
on random data and check whether the characteristics of label proportions continues to hold. 
It would also be interesting to examine how general these aspects of denoted label proportions are on real-world datasets, not only in MLPs but in related models that include deep layers, such as LUTs \citep{chatterjee2018learning}. 

\section{Conclusion}
\label{conclusion}

We have presented a framework and set of tools for examining denotation in the activations of neural networks, and we have taken initial steps in searching 
empirically for examples of such denotation on controlled datasets. Our experiments demonstrate that 
meanings of important auxiliary concepts are captured in a neural network, sometimes by individual neurons. 
Moreover, our results with convolutional networks show that a property can be captured by the structure of neural activations even if those activations do not denote the property linearly. We are unlikely to uncover such structure manually. Finally, we have shown that important quantitative characteristics of denoted properties,
such as label proportions, are dependent on neural position in a deep network, and we have suggested 
interpretations of these dependencies, as well as avenues for further research. 

There is much more to explore. The performance results we achieved with more sophisticated models suggests the presence of deep patterns of denotation; can we find chains of linear denotation for these properties explicitly? Can we infer salient properties learned by a network without a priori identification of those properties, solely by analyzing activation patterns? Can we empirically
validate our interpretation of observed label proportion of denoted properties in an object network? 
We look forward to exploring these problems in continuing work. 

\section{Broader impacts}
Understanding whether properties in a dataset are predictable from neural activations has theoretical interest: It can help us to better understand the structure of neural networks. But such understanding also has important, immediate application: These same techniques provide us with a tool that can be used to guard against implicit bias in predictive models. For example, suppose we are training a model to filter loan applications. Although the loan applications themselves might include information such as the gender and race of the applicant, we remove this information from the feature set in an attempt to avoid implicit bias. However, the model might still learn to discriminate based on proxies for these features in the remaining dataset. Because we have the information on these protected features in the original dataset, we can train an observer model on our model, using the protected features as labels. If the performance of our observer model is high, we can infer that the object model has indeed learned to discriminate by proxy, and we can work to eliminate proxies from the remaining features (or modify the object model to control for them). 

\section{Acknowledgements}
We would like to thank Danny Jachowski, Daniel Mahler, Dennis Michalopoulos, and J. Rafael Tena for helpful comments and suggestions on earlier drafts of this work. 

\bibliography{neural_denotation}

\end{document}